\documentclass{sig-alternate-05-2015}

\begin{document}

% Copyright
\setcopyright{acmcopyright}
%\setcopyright{acmlicensed}
%\setcopyright{rightsretained}
%\setcopyright{usgov}
%\setcopyright{usgovmixed}
%\setcopyright{cagov}
%\setcopyright{cagovmixed}

% DOI
\doi{xxxxxx}

% ISBN
\isbn{xxxxxx}

%Conference
\conferenceinfo{GECCO '16}{July 20--24, 2016, Denver, CO, USA}

\acmPrice{\$xx.xx}

%
% --- Author Metadata here ---
\conferenceinfo{GECCO}{'16 Denver, CO, USA}
%\CopyrightYear{2007} % Allows default copyright year (20XX) to be over-ridden - IF NEED BE.
%\crdata{0-12345-67-8/90/01}  % Allows default copyright data (0-89791-88-6/97/05) to be over-ridden - IF NEED BE.
% --- End of Author Metadata ---

\title{Evaluation of a Tree-based Pipeline Optimization Tool\\for Automating Data Science}

\numberofauthors{4} %  in this sample file, there are a *total*
\author{
% 1st. author
\alignauthor
Randal S.~Olson\\
       \affaddr{University of Pennsylvania}\\
       \affaddr{3700 Hamilton Walk}\\
       \affaddr{Philadelphia, PA 19104}\\
       \email{olsonran@upenn.edu}
% 2nd. author
\alignauthor
Nathan Bartley\\
       \affaddr{University of Chicago}\\
       \affaddr{5801 S. Ellis Avenue}\\
       \affaddr{Chicago, IL 60637}\\
       \email{bartleyn@uchicago.edu}
\and  % use '\and' if you need 'another row' of author names
% 3rd. author
\alignauthor
Ryan J.~Urbanowicz\\
       \affaddr{University of Pennsylvania}\\
       \affaddr{3700 Hamilton Walk}\\
       \affaddr{Philadelphia, PA 19104}\\
       \email{ryanurb@upenn.edu}
% 4th. author
\alignauthor
Jason H.~Moore\\
       \affaddr{University of Pennsylvania}\\
       \affaddr{3700 Hamilton Walk}\\
       \affaddr{Philadelphia, PA 19104}\\
       \email{jhmoore@upenn.edu}
}

% There's nothing stopping you putting the seventh, eighth, etc.
% author on the opening page (as the 'third row') but we ask,
% for aesthetic reasons that you place these 'additional authors'
% in the \additional authors block, viz.
%\additionalauthors{Additional authors: John Smith (The Th{\o}rv{\"a}ld Group,
%email: {\texttt{jsmith@affiliation.org}}) and Julius P.~Kumquat
%(The Kumquat Consortium, email: {\texttt{jpkumquat@consortium.net}}).}
% Just remember to make sure that the TOTAL number of authors
% is the number that will appear on the first page PLUS the
% number that will appear in the \additionalauthors section.

\maketitle
\begin{abstract}

As the field of data science continues to grow, there will be an ever-increasing demand for tools that make machine learning accessible to non-experts. In this paper, we introduce the concept of tree-based pipeline optimization for automating one of the most tedious parts of machine learning---pipeline design. We implement an open source Tree-based Pipeline Optimization Tool (TPOT) in Python and demonstrate its effectiveness on a series of simulated and real-world benchmark data sets. In particular, we show that TPOT can design machine learning pipelines that provide a significant improvement over a basic machine learning analysis while requiring little to no input nor prior knowledge from the user. We also address the tendency for TPOT to design overly complex pipelines by integrating Pareto optimization, which produces compact pipelines without sacrificing classification accuracy. As such, this work represents an important step toward fully automating machine learning pipeline design.\newline

\end{abstract}




%
%  Use this command to print the description
%


\begin{figure*}[t]
\begin{center}
\includegraphics[width=0.75\textwidth]{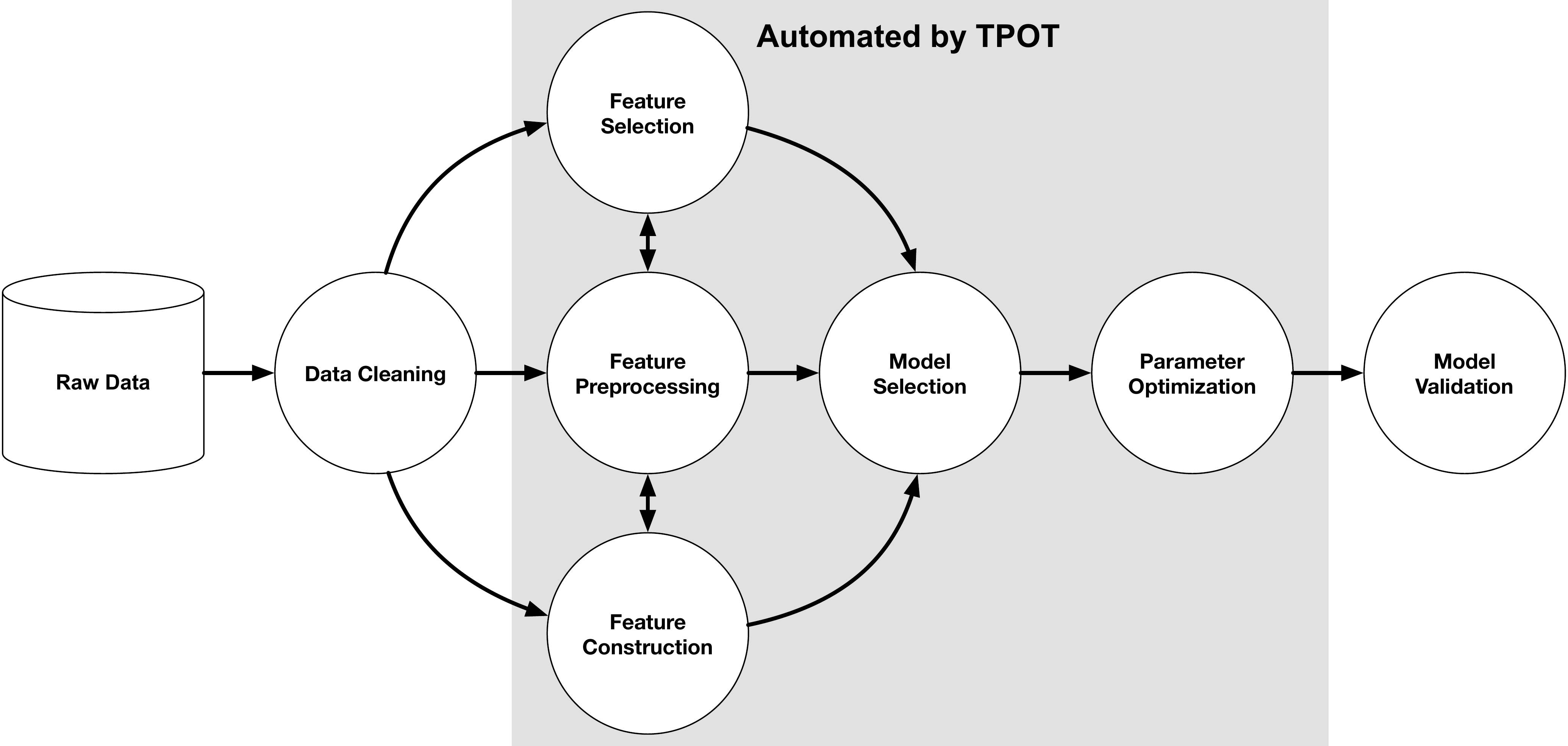}
\end{center}
\caption{A depiction of the typical supervised machine learning process. Before fitting a model of the data, the practitioner must prepare the data for modeling by performing an initial exploratory analysis (e.g., looking for missing or mislabeled data) and either correct or remove the offending records (i.e., data cleaning). Next, the practitioner may transform the data in some way to make it more suitable for modeling, e.g., by normalizing the features (i.e., feature preprocessing), removing features that are not useful for modeling (i.e., feature selection), and/or creating new features from the existing data (i.e., feature construction). Afterward, the practitioner must select a machine learning model to fit to the data (i.e., model selection) and choose the model parameters that allow the model to make the most accurate classification from the data (i.e., parameter optimization). Lastly, the practitioner must validate the model in some way to ensure that the model's predictions generalize to data sets that it was not fitted on (i.e., model validation), for example, by testing the model's performance on a holdout data set that was excluded from the earlier phases of the pipeline. The light grey area indicates the steps in the pipeline that are automated by the Tree-based Pipeline Optimization Tool (TPOT).}
\label{fig:tpot-ml-pipeline-diagram}
\end{figure*}

\section{Introduction}

Between 2011 and 2015, the number of self-reported data scientists more than doubled~\cite{RJMetrics2015}. At the same time, machine learning has returned to the forefront of academia, business, and government as data scientists discover new applications for algorithms that automatically learn and create actionable insights from data. Owing to this growth, there has been a commensurate demand for off-the-shelf tools that make machine learning more accessible, scalable, and flexible such that they can be applied across a wide variety of domains by non-experts. Unfortunately, the effective application of many machine learning tools typically requires expert knowledge of the tool and the problem domain, knowledge of the assumptions involved in the analysis, and/or the use of exhaustive brute force techniques. Thus, harnessing many machine learning tools is often a costly endeavor---both in terms of time and computation.

For example, a typical data scientist will approach a machine learning problem as demonstrated in Figure~\ref{fig:tpot-ml-pipeline-diagram}. Each step presents dozens of possible choices to make: How should the data be preprocessed (e.g., subset the features via feature selection? denoise the data with PCA?), what model should be used to fit the data (e.g., random forest? support vector machine?), what are the ideal model parameters for learning (e.g., how many trees in a random forest?), and so on. Experienced data scientists typically have a good sense for promising starting points in a given problem domain, but inexperienced data scientists can easily spend most of their time exploring myriad pipeline configurations before settling on the best one.

Theoretically, manual design of machine learning pipelines should no longer be necessary. In recent years, we have witnessed the development of intelligent systems in the field of evolutionary computation that consistently surprise us with their capabilities. From space antenna design~\cite{Hornby2011} to software development~\cite{Fredericks2013} and debugging~\cite{Forrest2009} to the study of finite algebras~\cite{Spector2008}, evolutionary algorithms have outperformed humans in a variety of domains that were previously considered exclusive to humans. If intelligent systems have been proven capable in so many domains, then we must ask ourselves: Can evolutionary algorithms automate the design of machine learning pipelines?\newline

In this paper, we report on the most recent development of an evolutionary algorithm called the Tree-based Pipeline Optimization Tool (TPOT) that automatically designs and optimizes machine learning pipelines~\cite{Olson2016EvoBIO}. TPOT uses a version of genetic programming~\cite{Banzhaf1998} to automatically design and optimize a series of data transformations and machine learning models that maximize the classification accuracy for a given supervised learning data set. In the following sections, we demonstrate TPOT's capabilities across a series of benchmarks, including simulated genetic analysis data sets and nine benchmark data sets from the well-known UC Irvine Machine Learning Repository~\cite{Lichman2013}. In particular, we show that TPOT is capable of discovering pipelines that achieve competitive performance by combining pre-existing algorithms in novel ways. We also compare the standard version of TPOT to {\em TPOT-Pareto}, a version of TPOT with Pareto optimization, and show that simultaneously optimizing pipeline accuracy and pipeline complexity leads to effective {\em and} compact pipelines.

\section{Related Work}

Historically, machine learning automation research has primarily focused on optimizing subsets of the pipeline~\cite{Hutter2015}. For example, grid search is the most commonly-used form of hyperparameter optimization that applies brute force search to explore a broad range of model parameters in order to discover the parameter set that allows for the best model fit. Recent research has shown that randomly evaluating parameter sets within the grid search often discovers the ideal parameter set more efficiently than exhaustive search~\cite{Bergstra2012}, which shows promise for intelligent search in the hyperparameter space. Bayesian optimization of model hyperparameters, in particular, has been effective in this realm and has even outperformed manual hyperparameter tuning by expert practitioners~\cite{Snoek2012}.

Another focus of machine learning automation research has been feature construction. One recent example of automated feature construction is the ``Data Science Machine,'' which automatically constructs features from relational dat\-abases via deep feature synthesis~\cite{Kanter2015}. In their work, Kanter {\em et al.} demonstrated the crucial role of automated feature construction in machine learning pipelines by entering their Data Science Machine in three machine learning competitions and achieving expert-level performance in all of them.

\begin{figure*}[t]
\begin{center}
\includegraphics[width=0.75\textwidth]{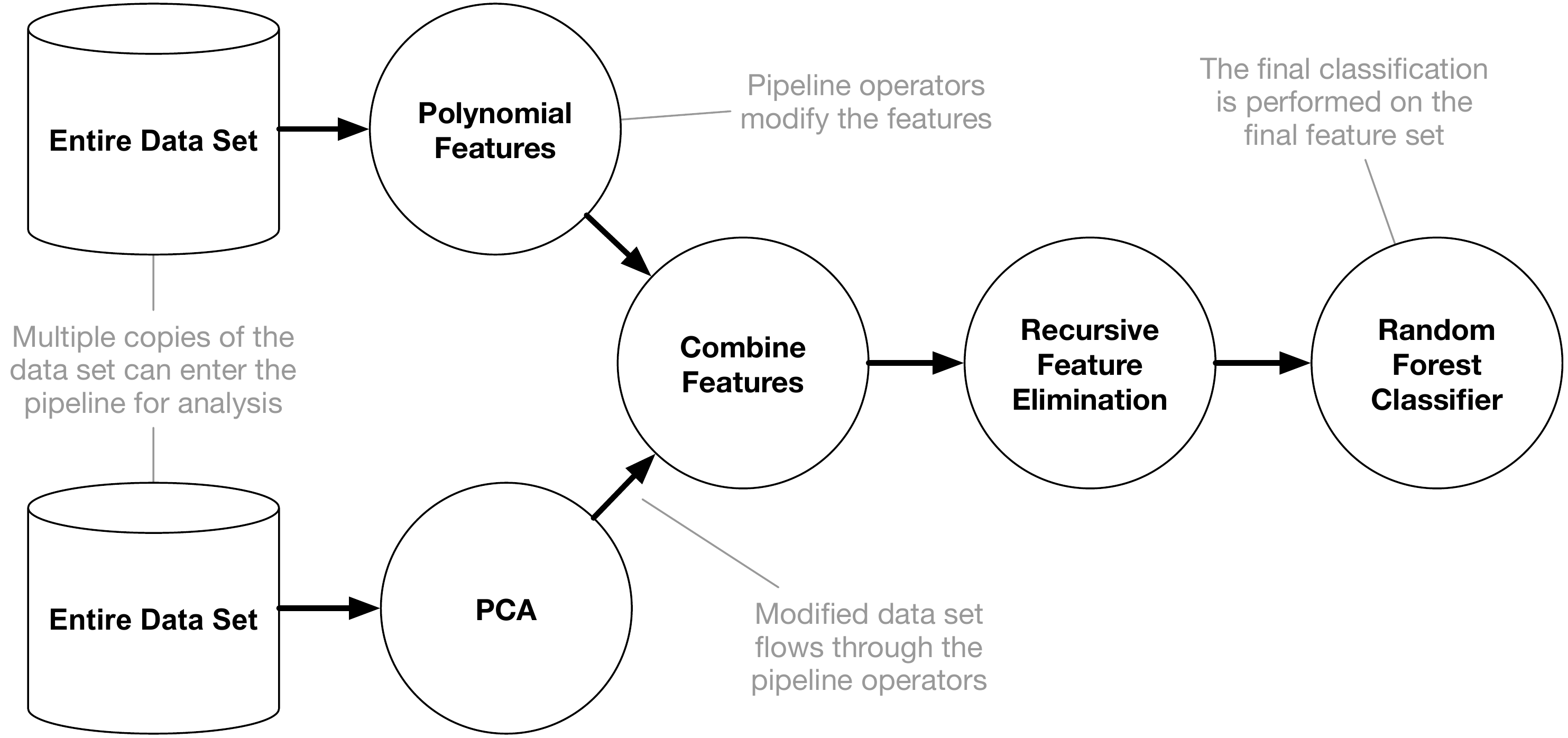}
\end{center}
\caption{An example tree-based machine learning pipeline. The data set flows through the pipeline operators, which add, remove, or modify the features in a successive manner. Combination operators allow separate copies of the data set to be combined, which can then be provided to a classifier to make the final classification.}
\label{fig:tpot-pipeline-example}
\end{figure*}

More recently, Fuerer {\em et al.} developed a machine learning pipeline automation system called auto-sklearn~\cite{Fuerer2015}, which uses Bayesian optimization to discover the ideal combination of feature preprocessors, models, and model hyperparameters to maximize classification accuracy. However, auto-sklearn explores a fixed set of pipelines that only include one data preprocessor, one feature preprocessor, and one model. Thus, auto-sklearn is incapable of producing arbitrarily large pipelines, which may be important for some machine learning analyses.

All of these findings point to one take-away message: Intelligent systems are capable of automatically designing portions of machine learning pipelines, which can make machine learning more accessible and save practitioners considerable amounts of time by automating one of the most laborious parts of machine learning. As such, the work presented in this paper establishes a blueprint for future research on the automation of machine learning pipeline design.

\section{Methods}

In this section, we describe tree-based pipeline optimization in detail, including the tools and concepts that underlie the Tree-based Pipeline Optimization Tool (TPOT). We begin this section by listing the basic pipeline operators that are currently implemented in TPOT. Next we describe how the operators are combined together into a tree-based pipeline, and then illustrate how tree-based pipelines can be evolved via genetic programming. Finally, we end this section by providing an overview of the data sets that we use to evaluate TPOT.

\subsection{Pipeline Operators}

Here we list the four main types of pipeline operators that are currently implemented in TPOT. All pipeline operators make use of existing implementations in scikit-learn~\cite{scikit-learn}. For further reading on these operators, refer to the scikit-learn online documentation and~\cite{MachineLearningBook}.

{\bf Preprocessors.} We implemented a standard scaling operator that uses the sample mean and variance to scale the features (StandardScaler), a robust scaling operator that uses the sample median and inter-quartile range to scale the features (RobustScaler), and an operator that generates interacting features via polynomial combinations of numerical features (PolynomialFeatures).

{\bf Decomposition.} We implemented RandomizedPCA, a variant of Principal Component Analysis that uses randomized SVD.

{\bf Feature Selection.} We implemented a recursive feature elimination strategy (RFE), a strategy that selects the top $k$ features (SelectKBest), a strategy that selects the top $n$ percentile of features (SelectPercentile), and a strategy that removes features that do not meet a minimum variance threshold (VarianceThreshold).

{\bf Models.} In this paper, we focus on supervised learning models. We implemented both individual and ensemble tree-based models (DecisionTreeClassifier, RandomForestClassifier, and GradientBoostingClassifier), non-probabilistic and probabilistic linear models (SVM and LogisticRegression), and $k$-nearest neighbors (KNeighborsClassifier).

\subsection{Assembling Tree-based Pipelines}

To combine all of these operators into a flexible pipeline structure, we implemented the pipelines as trees as shown in Figure~\ref{fig:tpot-pipeline-example}, with the different operators being nodes in the tree. Every tree-based pipeline begins with one or more copies of the input data set as the leaves of the tree, which is then fed into one of the four classes of pipeline operators: preprocessing, decomposition, feature selection, or modeling. As the data is passed up the tree, it is modified by that node's operator. When there are multiple copies of the data set being processed, it is possible to combine them into a single data set via a data set combination operator.

Each time a data set is passed through a modeling operator, the resulting classifications are stored such that the most recent classifier to process the data overrides any previous predictions, and the earlier classifier's predictions are stored as a new feature. Once the data set is fully processed by the pipeline (e.g., when the data set is passed through the Random Forest Classifier operator in Figure~\ref{fig:tpot-pipeline-example}), the final predictions are used to evaluate the overall classification performance of the pipeline. In all cases, we divide the data into stratified 75\% training and 25\% testing sets, such that the pipeline will only make predictions on and therefore only be evaluated on the testing set. This tree-based pipeline structure allows for arbitrary pipeline representations; for example, one pipeline could only apply operations in serial on a single copy of the data set, whereas another pipeline could just as easily work on several copies of the data set and combine them at the end before making a final classification.\newpage

\subsection{Evolving Tree-based Pipelines}

\begin{table}[t]
    \centering
    \caption{Genetic programming settings.}
    \begin{tabular}{l l}
        \hline \hline
        {\bf GP Parameter} & {\bf Value}\\ \hline
        Population size & 100\\
        Generations & 100\\
        Per-individual mutation rate & 90\%\\
        Per-individual crossover rate & 5\%\\
        TPOT selection & 10\% elitism,\\
         & rest 3-way tournament\\
         & (2-way parsimony)\\
        TPOT-Pareto selection & 5 copies of top 20\%\\
         & according to NSGA-II\\
        Mutation & Point, insert, \& shrink\\
         & 1/3 chance of each\\
        Unique replicate runs & 30\\
        \hline
    \end{tabular}
    \label{table:gp-settings}
\end{table}

To automatically generate and optimize these tree-based pipelines, we use a well-known evolutionary computation technique called genetic programming (GP) as implemented in the Python package DEAP~\cite{DEAP}. Traditionally, GP builds trees of mathematical functions to optimize toward a given criteria. In TPOT, we use GP to evolve the sequence of pipeline operators as well as each operator's parameters (e.g., the number of trees in a random forest or the number of feature pairs to select during feature selection) to maximize the classification accuracy of the pipeline. We follow a standard GP procedure with the settings described in Table~\ref{table:gp-settings}, where changes to the pipeline can modify, remove, or insert new sequences of pipeline operators into the tree-based pipeline.

In this paper, TPOT pipelines are evaluated based on their classification accuracy on the testing set. Here we also introduce an extension of TPOT, {\em TPOT-Pareto}, which uses Pareto optimization to optimize two separate objectives: maximizing the final classification accuracy of the pipeline as well as minimizing the pipeline's overall complexity (i.e., total number of pipeline operators). Since there is not a globally optimal solution that maximally optimizes both criteria, we maintain a Pareto front in TPOT-Pareto and select the pipelines for reproduction according to the NSGA-II selection strategy~\cite{Deb2002}

Through consecutive generations of evolution, TPOT's GP algorithm will tinker with the pipelines---adding new pipeline operators that improve fitness and removing redundant or detrimental operators---in an intelligent, guided search for high-performing pipelines. At the end of every TPOT run, we use the single best-performing pipeline ever discovered during the TPOT run (according to classification accuracy) as the representative pipeline.

\begin{figure*}
\begin{center}
\includegraphics[width=\textwidth]{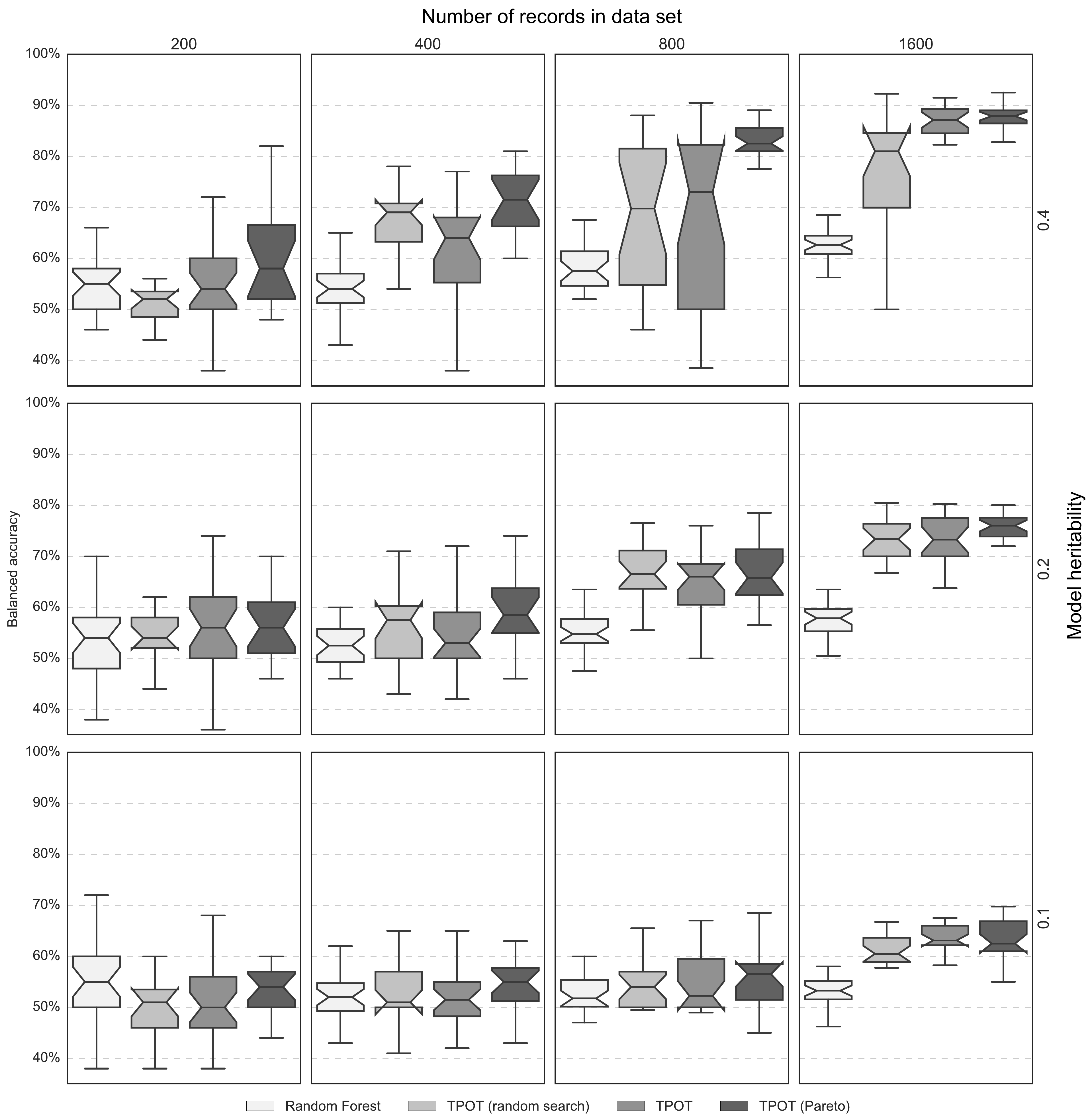}
\end{center}
\caption{Tree-based Pipeline Optimization Tool (TPOT) performance comparison across a range of data set sizes and difficulties. Each subplot on the grid shows the distribution of balanced cross validation accuracies on the holdout set (25\% of the data set), where each notched box plot represents a sample of 30 data sets. (Note: The notches in the box plots indicate 95\% confidence intervals of the median.) The experiments compared include a random forest with 500 decision trees (``Random Forest''), a version of TPOT with random generation of pipelines (``TPOT (random search)''), TPOT with guided search (``TPOT''), and a version of TPOT that uses Pareto optimization (``TPOT (Pareto)''). The subplots correspond to varying GAMETES configurations, where the x-axis modifies the number of records in the data set and the y-axis modifies the heritability in the model (where higher heritability reduces the amount of noise and vice versa). The grid ranges from easy configurations in the top right (larger data sets generated from higher heritability models) to difficult configurations in the bottom left (smaller data sets generated from low heritability models).}
\label{fig:tpot-gametes-comparison}
\end{figure*}

\begin{figure*}
\begin{center}
\includegraphics[width=\textwidth]{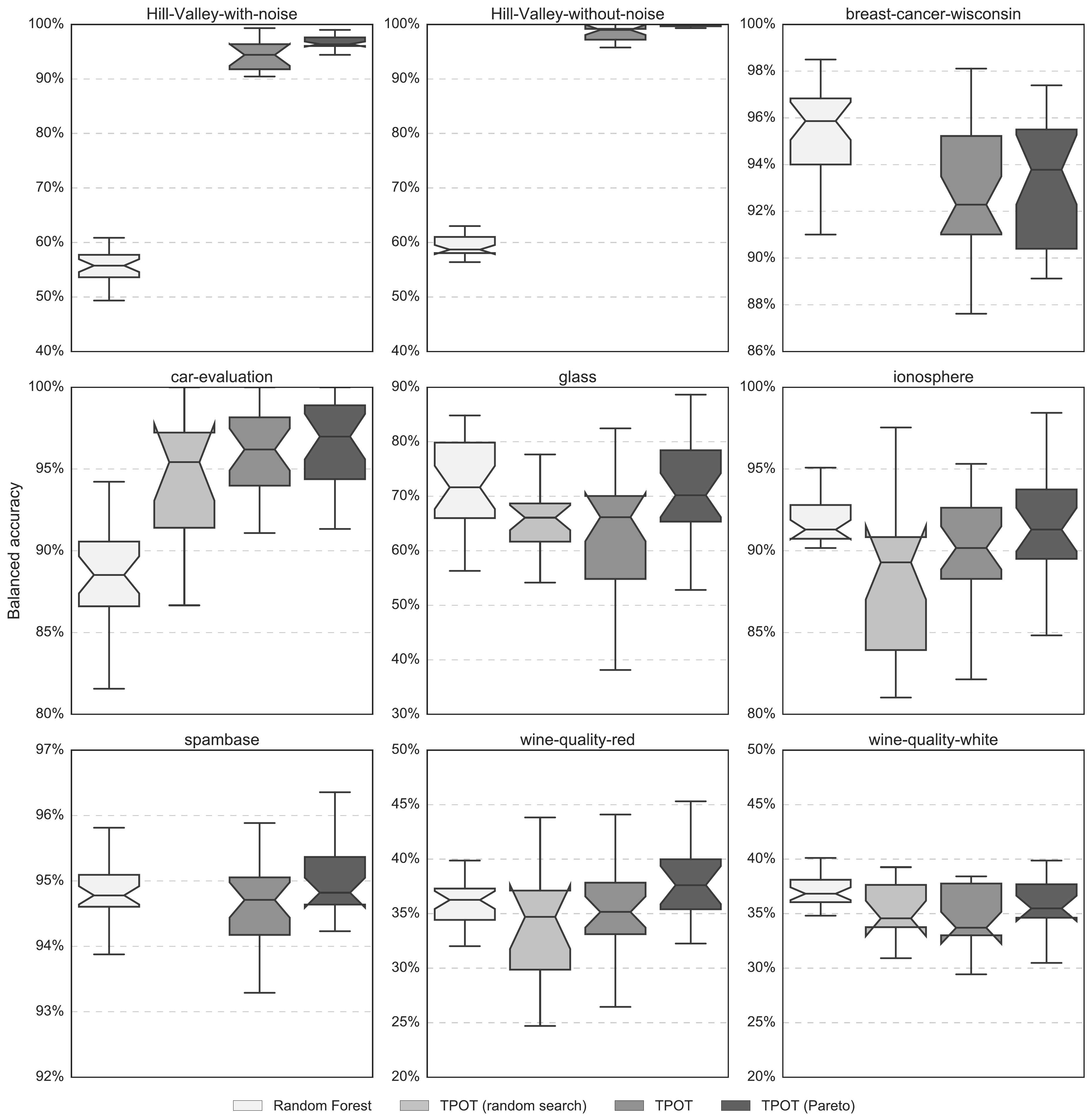}
\end{center}
\caption{Tree-based Pipeline Optimization Tool (TPOT) performance comparison across a series of UCI benchmark data sets. Each subplot on the grid shows the distribution of balanced cross validation accuracies on the holdout set (25\% of the data set), where each notched box plot represents a sample of 30 different cross validation divisions of the data set. (Note: The notches in the box plots indicate 95\% confidence intervals of the median.) The experiments compared include a random forest with 500 decision trees (``Random Forest''), a version of TPOT with random generation of pipelines (``TPOT (random search)''), TPOT with guided search (``TPOT''), and a version of TPOT that uses Pareto optimization (``TPOT (Pareto)''). The subplots correspond to different benchmark data sets. Note that some box plots for TPOT (random search) are missing because none of the replicates finished within 120 hours.}
\label{fig:tpot-uci-comparison}
\end{figure*}

\subsection{GAMETES Simulated Data Sets}

To evaluate TPOT, we adopt a diverse, complex simulation study design. We generate a total of 12 genetic models and 360 associated data sets using GAMETES~\cite{Urbanowicz2012a}, an open source software package designed to generate a diverse spectrum of pure, strict epistatic genetic models. GAMETES generates random, biallelic, {\em n}-locus single nucleotide polymorphism (SNP) models with ``pure'' epistasis, where all {\em n} loci, but no fewer, are predictive of disease status. We precisely generate these genetic models with specific heritabilities, SNP minor allele frequencies, and population prevalences.

In this paper, all data sets included 100 SNP attributes: 8 SNPs that are predictive of a binary case/control endpoint, and 92 SNPs that are randomly generated using an allele frequency between 0.05 and 0.5. The 8 predictive SNPs are simulated as four separate purely epistatic models, additively combined using the newly-added ``hierarchical'' data simulation feature in GAMETES. In doing so, each separate interaction model additively contributes to the determination of the endpoint, but the overall data set does not include main effects, i.e., direct associations between single SNP variables and the endpoint.

We simulate two-locus epistatic genetic models with heritabilities of (0.1, 0.2, or 0.4) and attribute minor allele frequencies of 0.2 in GAMETES and select the model with median difficulty from all those generated~\cite{Urbanowicz2012b}. From these models, we then generate data sets with a sample size of either 200, 400, 800, or 1600, within which each of the four underlying two-locus epistatic models carry an equal additive weight. For each model, we generate 30 replicate data sets, yielding a total of 360 data sets (i.e., 3 heritabilities * 4 sample sizes * 30 replicates). Together, this simulation study design allows us to evaluate TPOT across a broad range of data sets with varying difficulties and sample sizes to explore the limits of TPOT's modeling capabilities.

\subsection{UCI Benchmark Data Sets}

To further demonstrate TPOT's capabilities, we evaluate it on 9 hand-picked benchmark supervised learning data sets from the well-known UC-Irvine Machine Learning Repository~\cite{Lichman2013}. The purpose of these benchmarks is to demonstrate TPOT's performance across a broad range of application domains and data set types, as TPOT is intended to be a general-purpose supervised machine learning tool. For more information on these data sets, refer to their documentation at~\cite{Lichman2013}. The benchmark data sets are:

\begin{itemize}
    \item {\em Hill-Valley}: Two simulated data sets, with and without noise. Each record represents 100 points on a two-dimensional graph, where the algorithm must classify the series as either a Hill (a ``bump'' in the terrain) or a Valley (a ``dip'' in the terrain).
    \item {\em breast-cancer-wisconsin}: Data set containing continuous measurements from tumors. The algorithm must classify the tumor as benign or malignant.
    \item {\em car-evaluation}: Simulated data set containing categorical variables about cars. The algorithm must assign one of four classes indicating the car's acceptability for purchase.
    \item {\em glass}: Data set containing continuous measurements from various types of glass. The algorithm must assign one of seven classes indicating the glass type.
    \item {\em ionosphere}: Data set containing continuous measurements from high-frequency antennas. The algorithm must classify whether the signals are ``good'' or ``bad.''
    \item {\em spambase}: Data set containing word frequencies in e-mails. The algorithm must classify whether the e-mails are spam or not.
    \item {\em wine-quality}: Two data sets for red and white wine containing continuous measurements from various wines. The algorithm must classify the wine quality on a scale from 0 to 10 (11 classes).
\end{itemize}

\section{Results}

In this section, we compare TPOT's classification performance to that of two controls. The first control, {\em RF}, is a random forest with 500 decision trees, which is meant to represent a basic machine learning analysis with a state-of-the-art model. The second control, {\em TPOT-Random}, is a version of TPOT where the same number of pipelines are randomly generated, which is meant to explore whether guided search is useful for pipeline optimization. In addition, we compare TPOT to {\em TPOT-Pareto}, which is a version of TPOT that uses Pareto optimization to discover high-performing pipelines with the smallest pipelines possible. In all cases, we divide the data sets into stratified 75\% training and 25\% holdout sets, where the performance reported here is the balanced accuracy on the holdout sets.

In Figure~\ref{fig:tpot-gametes-comparison}, we compare these four experiments across a range of GAMETES data sets. In general, we see that all of the experiments achieve higher classification accuracy with larger data sets and/or higher heritability (i.e., the less noise) in the genetic model, which is to be expected. Interestingly, even in the easiest case with sample size 1600 and a heritability of 0.4, RF is incapable of discovering the epistatic interactions between the features in the data set and achieves only 63\% accuracy on average. In contrast, all versions of TPOT are capable of achieving 80\%+ accuracy on the easiest GAMETES data sets, which indicates that they were able to discover the epistatic interactions in the data set using a combination of feature preprocessing and modeling. This finding demonstrates how TPOT adds value over a simple machine learning analysis with no feature preprocessing.

Furthermore, Figure~\ref{fig:tpot-gametes-comparison} shows that, in most cases, all versions of TPOT perform more-or-less the same on average on the GAMETES data sets. This result suggests that guided search may not be vital for the automated design of pipelines because random search generally performs as well as guided search. However, TPOT-Pareto tends to be much more {\em consistent} in discovering effective classification pipelines, as indicated by the lower variance in TPOT-Pareto's accuracy distributions---especially in the GAMETES data sets with larger sample sizes and higher heritability.

\begin{figure}[t]
\begin{center}
\includegraphics[width=0.4\textwidth]{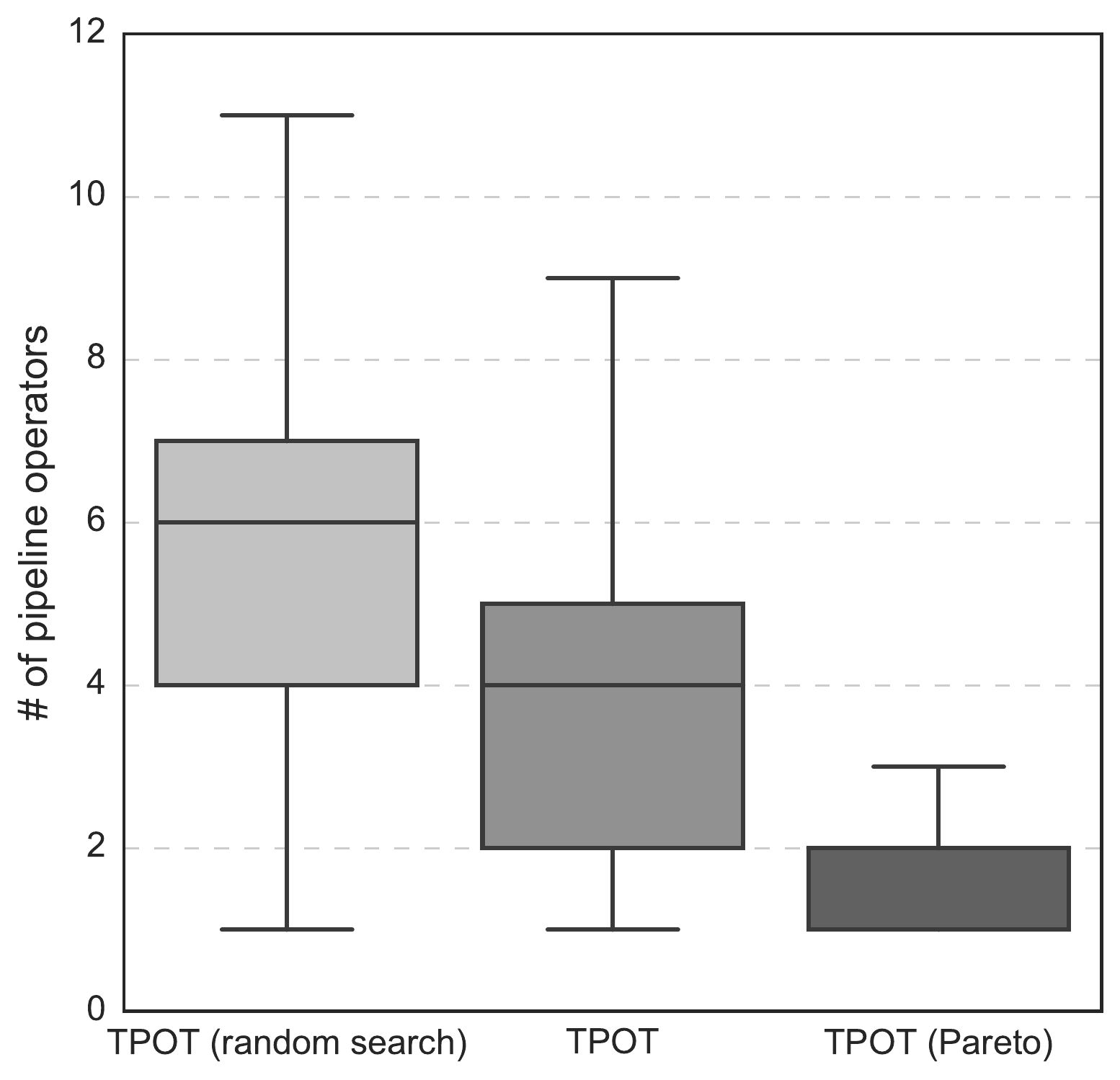}
\end{center}
\caption{Comparison of the final pipeline sizes across all of the TPOT experiments. The experiments compared include a version of TPOT with random generation of pipelines (``TPOT (random search)''), TPOT with guided search (``TPOT''), and a version of TPOT that uses Pareto optimization (``TPOT (Pareto)'').}
\label{fig:tpot-pipeline-size-comparison}
\end{figure}

Figure~\ref{fig:tpot-uci-comparison} compares the four experiments on the series of UCI benchmark data sets. Again, TPOT achieves the same classification accuracy as RF across most of these data sets, and achieves significantly higher classification accuracy in the {\em Hill-Valley} and {\em car-evaluation} data sets. With the {\em Hill-Valley-without-noise} data set in particular, TPOT-Pareto achieves 100\% accuracy in all 30 replicates, which outperforms even the standard version of TPOT. This finding again demonstrates the value of automated pipeline design that intelligently explores many different ways of preprocessing the data prior to modeling it.

Similar to the GAMETES data set comparisons, Figure~\ref{fig:tpot-uci-comparison} also shows that TPOT-Random typically performs as well as the versions of TPOT with guided search. However, there are some major drawbacks to randomly generating TPOT pipelines that are presented here. For one, randomly generating TPOT pipelines tends to be much slower than optimizing pipelines via guided search because some of the random pipelines are needlessly complex and take several hours to evaluate. As an example of these massive randomly generated pipelines, none of the TPOT-Random replicates running on the {\em Hill-Valley} and {\em spambase} data sets (the larger data sets) finished within 120 hours and had to be terminated prematurely. On the other hand, all TPOT and TPOT-Pareto replicates finished the same number of evaluations in less than 48 hours. Furthermore, even though TPOT-Random pipelines perform nearly as well as regular TPOT pipelines, TPOT-Random pipelines tend to be needlessly complex and contain 6 pipeline operators on average (Figure~\ref{fig:tpot-pipeline-size-comparison}). In contrast, TPOT and TPOT-Pareto can achieve the same performance with 4 and 2 operators on average, respectively. Thus, even though all versions of TPOT usually achieve the same accuracy, TPOT-Pareto in particular discovers pipelines that are significantly more compact.

\section{Discussion}

In this paper we have shown that, in many cases, automated machine learning pipeline design and optimization can provide a significant improvement over a basic machine learning analysis while requiring little to no input nor prior knowledge from the user. However, it is important to note that the goal of automated pipeline design is not to replace data scientists nor machine learning practitioners. Rather, we aim for the tree-based pipeline optimization tool (TPOT) to be a ``Data Science Assistant'' that explores the data, discovers novel features in the data, and recommends pipelines to the user. From there, the user is free to export the pipelines and integrate their domain knowledge as they see fit. To aid in this goal, we have released TPOT as an open source Python package that provides a flexible implementation of the concepts introduced in this paper. We encourage interested practitioners to involve themselves in the project on GitHub (http://github.com/rhiever/tpot).

One challenge raised in this paper is that TPOT with randomly generated pipelines consistently achieves the same accuracy as versions of TPOT with guided search (Figures~\ref{fig:tpot-gametes-comparison} \&~\ref{fig:tpot-uci-comparison}). In practice, accuracy is not the only criteria by which a pipeline is evaluated. For one, the randomly generated pipelines tend to be slower than the pipelines generated through guided search, as shown by the inability for TPOT-Random to evaluate 10,000 pipelines (100 population size x 100 generations) within a 120-hour period for several of the data sets (Figure~\ref{fig:tpot-uci-comparison}). As a consequence, randomly generating pipelines will quickly become computationally infeasible as data sets grow beyond 2,000 records. Furthermore, due to the compactness of the TPOT and TPOT-Pareto pipelines (Figure~\ref{fig:tpot-pipeline-size-comparison}), these pipelines are much easier for a user to interpret and apply in a production setting. Thus, guided evolutionary search plays a vital role in the automated design of machine learning pipelines that cannot be captured by random search.

Of course, TPOT is still in its early stages of development and there are still many improvements to be made. In particular, TPOT is still fairly slow on large data sets and often requires several hours (if not days) to properly analyze a large data set. In the near future, we plan to explore the use of auto-sklearn~\cite{Fuerer2015} and other heuristics to seed the TPOT population with promising pipelines, which can ``kick start'' the TPOT population. Similarly, we plan to integrate a learning system into the genetic programming mutation and crossover operators to bias these operators toward changes that tend to improve fitness, with the goal of spending less time exploring detrimental changes. By doing so, we hope to enable TPOT to deliver effective pipelines in a speedier manner.

\section{Conclusions}

Tree-based pipeline optimization is a new technique that shows significant promise for 1) making machine learning tools more accessible to non-experts and 2) saving practitioners considerable amounts of time by automating the most tedious parts of machine learning. In this paper, we demonstrated that TPOT achieves a similar level of performance as a basic machine learning analysis across a wide variety of data sets without any input nor prior knowledge from the user. Furthermore, in several cases, TPOT was able to automatically discover combinations of preprocessing and modeling operators that significantly outperformed a basic machine learning analysis. Finally, by integrating Pareto optimization into TPOT, we demonstrated that TPOT can design compact, easy-to-interpret pipelines without sacrificing classification accuracy. As such, this work represents an important step toward fully automating machine learning pipeline design.

%\section{Acknowledgments}

%We thank the Penn Medicine Academic Computing Services for the use of their computing resources. This work was supported by National Institutes of Health grants LM009012, LM010098, and EY022300.

\bibliographystyle{abbrv}
\bibliography{references}
% ACM needs 'a single self-contained file'!
%
\end{document}